%% file: main.tex
\pdfoutput=1
\documentclass[11pt]{article}
\usepackage{acl}
\usepackage{times}
\usepackage{latexsym}
\usepackage{multicol}
\usepackage[T1]{fontenc}
\usepackage[utf8]{inputenc}
\usepackage{microtype}
\usepackage{inconsolata}
\usepackage{multirow}
\usepackage{graphicx}
\usepackage{pgfplots}
\pgfplotsset{width=10cm,compat=1.16}
\usetikzlibrary{patterns}
\title{Unified Examination of Entity Linking in Absence of Candidate Sets}

\author{Nicolas Ong \and Hassan S. Shavarani\and Anoop Sarkar\\
  School of Computing Science \\
  Simon Fraser University \\
  BC, Canada \\
  \texttt{\{nong, sshavara, anoop\}@sfu.ca} \\}

\begin{document}
\maketitle
\begin{abstract}
Despite remarkable strides made in the development of entity linking systems in recent years, a comprehensive comparative analysis of these systems using a unified framework is notably absent. This paper addresses this oversight by introducing a new black-box benchmark and conducting a comprehensive evaluation of all state-of-the-art entity linking methods. We use an ablation study to investigate the impact of candidate sets on the performance of entity linking. Our findings uncover exactly how much such entity linking systems depend on candidate sets, and how much this limits the general applicability of each system.  We present an alternative approach to candidate sets, demonstrating that leveraging the entire in-domain candidate set can serve as a viable substitute for certain models. We show the trade-off between less restrictive candidate sets, increased inference time and memory footprint for some models\footnote{\href{https://github.com/NicolasOng/gerbil_connects}{https://github.com/NicolasOng/gerbil\_connects} contains all the evaluated models as well as our evaluation source code.}.

\end{abstract}

\section{Introduction}
Entity linking\footnote{See Appendix \ref{appendix:a} for an in-depth review of the modules in the entity linking pipeline.} is the task of annotating plain text with references to pre-defined entries - also known as entities - in a knowledge base. The annotation process involves finding specific parts of the text - also known as mentions - that may contain a knowledge base reference, and linking those to the right knowledge base record. For example, in the headline \texttt{SOCCER - LATE GOALS GIVE \underline{JAPAN} WIN OVER SYRIA}, the entity linking system can link \texttt{JAPAN} (underlined) to \textit{Japan national football team}.

Candidate generation is an integral part of many entity linking systems, in which a small set of candidate knowledge base entities are selected for each potential candidate span. Many recent entity linking techniques assume the presence of the pre-existing candidate sets for all potential spans. The most popular such set is KB+Yago \cite{D17-1277} which is used in a number of recent methods \cite{kolitsas-etal-2018-end, peters2019knowledge, poerner2019bert, feng2022efficient, REL_EL, de2020autoregressive, kannan-ravi-etal-2021-cholan, shavarani2023spel}. Another less popular set is PPRforNED \cite{pershina-etal-2015-personalized} which assumes the test sets are available to the annotators. Due to this assumption, very few recent publications evaluate using PPRforNED \cite{martins-etal-2019-joint, de2020autoregressive, de2021highly, shavarani2023spel}.

Unified evaluation of the different entity linking systems with respect to the application of candidate sets should play a crucial role in a better understanding of the strengths and weaknesses of each system. This will give the research community and commercial deployments better ways to select the most suitable system based on their needs while providing them a platform to identify avenues for enhancement.

In this paper, we propose a novel unified black-box evaluation framework for recent state-of-the-art entity linking systems. Our contributions in this paper are as follows:

(1) We unify the evaluation setup for the systems using GERBIL \citep{GERBIL} and \texttt{gerbil\_connect} \cite{shavarani2023spel}, and black-box evaluate the systems over the same benchmark dataset CoNLL/AIDA \cite{D11-1072} which allows us to abstract away their internal model structure and decoding algorithms.

(2) We discuss the importance of the pre-built candidate sets for obtaining good results on benchmarks in entity linking. However, candidate sets are not always available, and the literature lacks a systematic evaluation of the entity linking systems in absence of the candidate sets. To fill this gap, we suggest an experimental setup to replace them with a feasible set; the entire in-domain vocabulary of the benchmark dataset. Please note that our focus in these experiments is not to re-implement each technique, but rather to evaluate the \textit{resilience} of the entity linking systems in absence of the carefully hand-crafted candidate sets.

(3) We examine the \textit{adaptability} of the entity linking systems in the literature to unseen test data using the novel \texttt{AIDA/testc} dataset \citep{shavarani2023spel} which contains new annotations on news stories in 2020 with 924 novel entities.

\section{Unified Black-Box Evaluation}
We benchmark the recent entity linking systems, unchanged and as provided originally by their authors. In these experiments, we intend to examine the suitability of these systems as off-the-shelf systems which can be integrated in future downstream applications. 

We unify the evaluation environment as GERBIL \cite{GERBIL} plus \texttt{gerbil\_connect} \cite{shavarani2023spel}. In the evaluation procedure, GERBIL will provide the testing documents to \texttt{gerbil\_connect} and receives the entity annotations in the format of (\textit{begin char.}, \textit{end char.}, \textit{entity annotation}) from \texttt{gerbil\_connect}. We implement \texttt{gerbil\_connect} tailored to each entity linking system so that it can transform the evaluation documents to readable inputs for each system. Specifically, we have (1) utilized NLTK's word tokenizer\footnote{\href{https://www.nltk.org/api/nltk.tokenize.html}{https://www.nltk.org/api/nltk.tokenize.html}} to transform raw non-tokenized evaluation sets into their expected CoNLL tokenized format for the models that depend on reading from AIDA test files \citep{peters2019knowledge, poerner2019bert, feng2022efficient}, (2) simulated long text splitting and result merging strategies for the models with input length constraints \citep{peters2019knowledge, poerner2019bert, feng2022efficient, de2020autoregressive}, (3) implemented a subword token id to character id conversion for the models that output annotations as tokenized subword ids \citep{peters2019knowledge, poerner2019bert, feng2022efficient}, and (4) provided the external data sources such as the pre-built candidate sets to the model initializers where necessary \citep{de2020autoregressive, de2021highly}. Appendix \ref{appendix:b} provides more details on the mentioned alterations.
Empirically, running the models without adding these techniques significantly hurts performance. Removing the tokenization  step alone can drop the model performance by up to 20 Micro-F1 points.

Once done, \texttt{gerbil\_connect} translates the produced annotations in each system back to the unified annotation format, understandable for GERBIL. We train the models that are not released by the authors \citep{poerner2019bert, feng2022efficient}, using their own released source code, and do not consider the models which we were not able to acquire their training source code or were not able to get their training scripts to converge. \citep{martins-etal-2019-joint, Fevry, mrini-etal-2022-detection, kannan-ravi-etal-2021-cholan, broscheit2020investigating, xiao2023instructed}
Appendix \ref{appendix:c} briefly discusses each of the evaluated systems. We direct the readers to the original publication of each model for a more in-depth understanding of their contributions.

We use CoNLL/AIDA evaluation sets \texttt{testa} and \texttt{testb} - reported by all entity linking systems tested in different evaluation frameworks - as well as the newly annotated AIDA/\texttt{testc} evaluation set. The results tables show the GERBIL InKB Micro-F1 evaluation results.

\begin{table*}[ht!]
    \centering
    \begin{tabular}{l|ccc|cc}
        \multirow{2}{*}{~} & \multicolumn{3}{c}{Micro-F1} & \multicolumn{2}{|c}{Difference}
        \\\cline{2-6}
        & \texttt{testa} & \texttt{testb} & \texttt{testc} & \texttt{testa} & \texttt{testb}     \\
        \hline\hline
        \citet{kolitsas-etal-2018-end}   & 89.50  & 82.44 & 65.75 & +0.10 & +0.04 \\
        \citet{peters2019knowledge} &&&&&\\
        \ \ \ \ KnowBert-Wiki    & 76.74  & 71.68 & 54.12 & -3.46 & -2.72 \\
        \ \ \ \ KnowBert-W+W     & 77.19  & 71.69 & 53.92 & -4.91 & -2.01 \\        
        \citet{poerner2019bert}          & 89.40  & 84.83 &  65.93 & -1.40 & -0.17 \\
        \citet{REL_EL}&&&&&\\                 
        \ \ \ \ Wiki 2014        & 83.30  & 82.53 &  71.69 & - & -0.77                  \\
        \ \ \ \ Wiki 2019        & 79.64  & 80.10 & 73.54 & - & -0.40 \\
        
        \citet{de2020autoregressive}     & 90.09  & 82.78 & 75.60 & - & -0.92 \\
        
        \citet{de2021highly}             & 87.29  & 85.65  & 47.54 & - & +0.15 \\
        
        \citet{zhang2021entqa}           & 86.81  & 84.30 & 72.55 & - & -1.50 \\

        \citet{feng2022efficient}        & 87.64  & 86.49 & 65.05 & - & +0.19 \\
        
        \citet{shavarani2023spel}&&&&&\\
        \ \ \ \ large-500K (no cnds.) & 89.72  & 82.25 & 77.54 & +0.02 & +0.05 \\
        \ \ \ \ large-500K (Kb+Yago)  & 89.89  & 82.88 & 59.50 & +0.09 & +0.08\\
        \ \ \ \ large-500K (PPRforNED)   & 91.58  & 85.22 & 46.98 & +0.08 & +0.02 \\
        \hline\hline
    \end{tabular}
    \caption{Comparison of recent entity linking systems within the unified black-box testing framework of GERBIL + \texttt{gerbil\_connect}. Difference column reports the difference between our unified evaluation environment and the originally reported numbers. We have assessed all models twice for consistency. Except for \citep{de2020autoregressive}, all models yielded identical scores, while \citet{de2020autoregressive} showed a low variance of 0.08 in the results. Thus, the results mirror those reported by GERBIL, with the exception of \citep{de2020autoregressive}, which is averaged over two runs.}
    \label{tab:el_baselines_results}
\end{table*}

Table \ref{tab:el_baselines_results} presents the unified black-box evaluation results. The necessary unification adjustments mentioned above and the evaluation format has caused some evaluation scores to deviate from their original reported results. However, we have tried to control for this as much as possible. The \texttt{Difference} columns in Table \ref{tab:el_baselines_results} reflects on the mentioned score deviations.

In our experiments, we found that \citep{peters2019knowledge} suffered the most, with an approximate loss of 5\% when comparing our results to the originally reported scores. The rest of the models were hit by at most 2\%, confirming the reliability of our framework for further analysis. \texttt{testc} is a more challenging evaluation set which contains novel entities that typically hurt model recall. In our experimental results, we found that text generation models (from previous work) performed better on \texttt{testc}. However, the best performing model was not generative, but rather used structured prediction \cite{shavarani2023spel}.

\section{Candidate Set Ablations}

Candidate sets are an integral part of entity linking systems, many of which assume the presence of good quality sets to perform well. Although this assumption holds when linking to English Wikipedia, it does not necessarily hold when considering other ontologies (e.g. UMLS; \citealp{UMLS}) and languages\footnote{See \citet{2020.emnlp-main.630} for more discussion.}. 

In this section, we ablate the mention-specific candidate sets from the entity linking systems to study their performance in absence of the hand-crafted candidate sets. For our experiments, we selected the candidate-set-independent setting of the models in any system that provides such a setting. For the other systems that require a candidate set, and we cannot remove the candidate set dependence, we return the entire in-domain mention vocabulary of AIDA \citep[the in-domain fixed candidate set of][]{shavarani2023spel} as the replacement for the required candidate sets (5598 entities including the \textit{None} entity). Where applicable, we add priors such that each candidate has an equal probability.

Table \ref{tab:el_no_candidate_results} demonstrates the evaluation results of the models with Micro-F1 scores above 1.0 after considering the candidate-independent version of the models, or the candidate set expansion.

\begin{table}[ht!]
    \centering
    \addtolength{\tabcolsep}{-0.4em}
    \scalebox{0.95}{
    \begin{tabular}{l|l|ccc}
        \multicolumn{1}{l}{}&\multirow{2}{*}{~}               & \multicolumn{3}{c}{Micro-F1} \\\cline{3-5}
        \multicolumn{1}{l}{} &                                 & \texttt{testa} & \texttt{testb} & \texttt{testc}     \\
        \hline\hline
        
        \multirow{4}{*}{a)}&\citet{de2020autoregressive}     & 85.15  & 78.98 & 75.62 \\ 
        
        &\citet{de2021highly}             & 62.00  & 49.51 & 37.05 \\
        
        &\citet{zhang2021entqa}           & 86.81  & 84.30 & 72.55 \\
        &\citet{shavarani2023spel} & 89.72  & 82.25 & 77.54 \\
        \hline
        \multirow{2}{*}{b)} &\citet{poerner2019bert}          & 22.81  & 18.81 &  17.56 \\
        
        &\citet{feng2022efficient}        & 35.00  & 32.58 & 27.48 \\
        \hline\hline
    \end{tabular}
    }
    \caption{Comparison of entity linking systems after a) running the model with no access to hand-crafted candidate sets b) modifying the model to consider the entire AIDA in-domain vocabulary as the candidate set.}
    \label{tab:el_no_candidate_results}
\end{table}

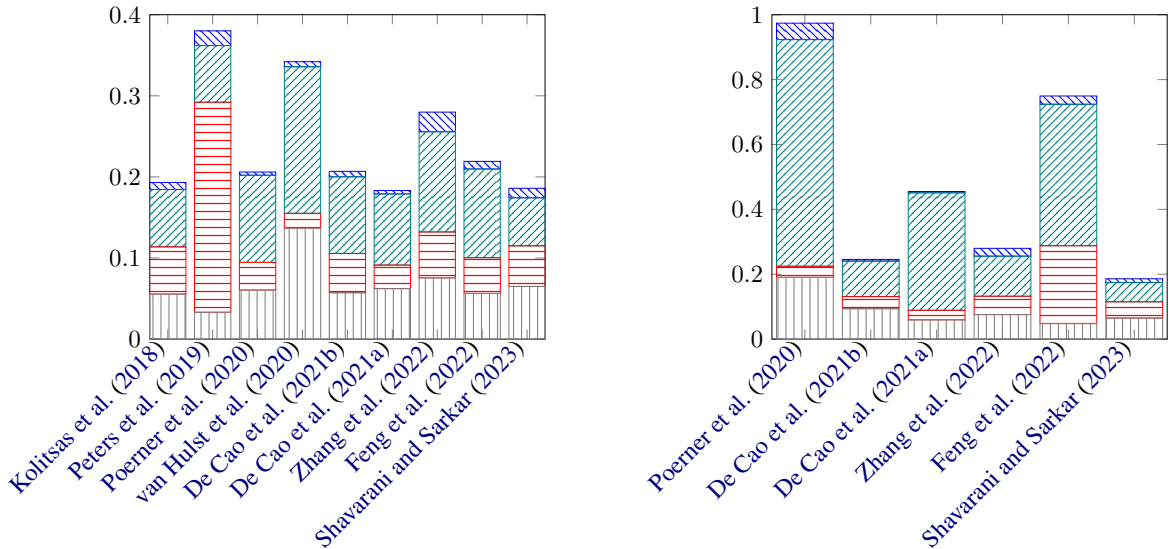
\begin{figure*}[ht]
    \centering
    \begin{multicols}{2}
        \scalebox{0.85}{\input{testa_before_analysis_figure}}
    \columnbreak
        \scalebox{0.85}{\input{testa_after_analysis}}
    \end{multicols}
    \caption{Entity linking error distribution in four categories of over-generated (gray, vertical), under-generated (red, horizontal), incorrect entity (teal, north east) and incorrect mention (blue, north west) before candidate set ablations (left) and after the ablations (right). The y-axis is the error analysis ratio as described below.}
    \label{fig:testa_error_analysis}
\end{figure*}

We experimented with removing candidate sets altogether, but the models that appear in Table \ref{tab:el_baselines_results} but do not appear in Table \ref{tab:el_no_candidate_results} essentially failed without candidate sets, resulting in Micro-F1 scores close to 0. These results demonstrate that most entity linking systems are too intertwined with their candidate sets and without this additional data resource, the systems do not produce useful results and are too brittle to be used in real-world production deployments.

Table \ref{tab:el_no_candidate_results} results prove that generation-based systems are more \textit{resilient} against candidate sets. Nonetheless, without given candidate sets, \citet{de2020autoregressive} and \citet{de2021highly} lose approximately 5\% and 20-30\% of their best performance, respectively.
\citet{shavarani2023spel} - a non-generation-based system, designed without dependence on candidate sets and only using these resources to improve performance - suffers the least and loses only 2\% of its best performance without candidate sets.

The largest performance drop in our experiments correlates with using mention-entity similarity methods for entity disambiguation, where a representation of the mention and entity are compared to determine the most relevant entity.

In these systems, models that generate mention representations by combining candidate entity representations see their performance decreased to 20\%-35\%, while models that generate mention representations by combining the word or token representations within or surrounding the mention perform too poorly to be present in Table \ref{tab:el_no_candidate_results}.

\citet{shavarani2023spel} and \citet{de2020autoregressive} only show an approximate 2\% drop in performance, showing that they can easily handle a larger set of candidate entities.

The larger candidate sets lead to longer inference times. The run time for \citet{feng2022efficient, kolitsas-etal-2018-end, poerner2019bert, peters2019knowledge} that compare the mentions to each entity in the candidate set increases by 90x, 50x, 25x, and 10x, respectively. \citet{REL_EL} does not follow this trend since it selects the 30 candidate entities with the highest prior before performing entity disambiguation.

\textbf{Error Analysis.} We store the produced annotations from each system reported in Table \ref{tab:el_baselines_results} (w/ candidate sets) and Table \ref{tab:el_no_candidate_results} (w/o candidate sets),  and compare their produced annotations with the expected annotations of AIDA/\texttt{testa} (4791 annotations). For models with multiple reported settings, we select the setting correlated to AIDA/\texttt{testc} as it represents the most generalization-capable setting for unseen in-domain documents.

We count the number of annotations in four error categories of over-generated, under-generated, incorrect mention and incorrect entity, and divide each by the total number of gold annotations. Figure \ref{fig:testa_error_analysis} presents the calculated error analysis ratios. Over-generation refers to annotations made by the model and not in the gold set. Under-generation refers to annotations in the gold set but not made by the model. Incorrect entity refers to annotations where the model linked the wrong entity. Incorrect mention refers to annotations where the span's start or end is incorrect.

Before ablation of candidate sets (Figure \ref{fig:testa_error_analysis}-left), \citep{REL_EL} has the highest rate of over-generation followed by \citep{zhang2021entqa}, while \citep{peters2019knowledge} shows the lowest over-generation rate. On the other hand \citep{peters2019knowledge} has the highest under-generation ratio establishing itself as a conservative entity linking system.

Comparing the performance of entity systems w/ and w/o candidate sets, the biggest increase is seen in incorrect entity prediction ratios, confirming the dependence of entity linking systems to candidate sets. \cite{poerner2019bert} sees the biggest increase in incorrect entity predictions followed by \cite{feng2022efficient}. While \cite{zhang2021entqa} and \cite{shavarani2023spel} report the smallest rate increase in this category as these methods are less dependent on candidate sets.
\citep{feng2022efficient} on the other hand shows an increase in under-generation signaling the effect of candidate sets in prediction confidence for this system.

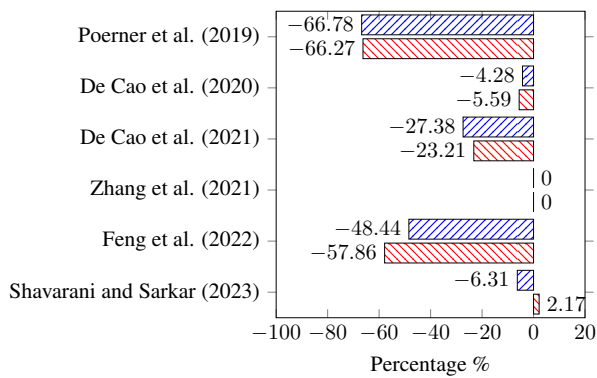
\begin{figure}[ht!]
    \centering
        \scalebox{0.75}{\input{testa_pr_diff_nocite}}
    \caption{Entity linking micro precision (blue, north east) and recall (red, north west) score differences over \texttt{testa} between model's original configuration and candidate set ablation configuration.}
    \label{fig:testa_pr_diff}
\end{figure}

Figure \ref{fig:testa_pr_diff} illustrates the disparities in precision and recall pre- and post-ablation of candidate sets for the models outlined in Table \ref{tab:el_no_candidate_results}. Our findings reveal that candidate sets significantly enhance precision and recall. With the exception of \cite{zhang2021entqa}, which generates candidates in real-time, the other systems show that without candidate sets there is a substantial decrease in precision and recall, exceeding 60\% for \cite{poerner2019bert}.

\section{Conclusion}
In this study, we have successfully established a unified black-box evaluation framework for modern entity linking techniques. We provide an in-depth ablation analysis to examine the significance of hand-crafted candidate sets for  the entity linking task. Our findings confirm that modern entity linking systems are excessively dependent on candidate sets. Our study shows that we need to be less reliant on hand-crafted candidate sets in order to ensure robust, versatile and accurate entity linking systems.

\bibliography{references}

\appendix

\section{A More In-Depth Literature Review}\label{appendix:a}

Entity linking can be viewed as a multi-task pipeline which performs mention detection, candidate generation, and entity disambiguation \citep[see][]{broscheit2020investigating, REL_EL, shavarani2023spel}. \textit{Mention detection} focuses on the discovery of potential references in the given text to knowledge base entities. \textit{Candidate Generation} short-lists a small set of candidate knowledge base entities for each potential span. \textit{Entity Disambiguation} selects one of the short-listed entities to link to the selected span.

\textbf{Mention Detection.} Common approaches include: (1) listing every possible span of length up to $n$ (usually 5) tokens \citep{kolitsas-etal-2018-end, peters2019knowledge, poerner2019bert, feng2022efficient}, (2) using a named entity recognition (NER) tool \citep{REL_EL, D11-1072} such as Flair \citep{akbik-etal-2018-contextual} or the Stanford NER Tagger \citep{finkel-etal-2005-incorporating}, (3) scoring each token as a potential mention and merging the consecutive predictions referring to the same entity \citep{broscheit2020investigating, shavarani2023spel}, and (4) probabilistic prediction of tokens beginning, ending, or outside (\texttt{BIO}) of mentions \citep{kannan-ravi-etal-2021-cholan, Fevry, de2021highly, zhang2021entqa, xiao2023instructed}.

\textbf{Candidate Generation.} This module selects the most relevant of all possible entities for a candidate span in the input text. As previously mentioned, most models rely on pre-existing candidate sets such as KB+Yago \cite{D17-1277}. \citet{kannan-ravi-etal-2021-cholan} use the BM25 \citep{robertson1995okapi} algorithm to rank entities for a mention. \citet{zhang2021entqa} and \citet{xiao2023instructed} explore document-level candidate sets that aid the mention detection step. Both rank the dot product scores of entity embeddings by an input document embedding. \citet{zhang2021entqa} use a fine-tuned BLINK \citep{wu-etal-2020-scalable} model. In the extreme case, some methods assume the entire entity vocabulary as the candidate set \citep{broscheit2020investigating, Fevry, shavarani2023spel}. Such methods usually limit the size of the entity vocabulary to 500K-700K of the most frequent entities. Wikipedia contains over 6.5M entities.
In candidate generation, mentions with an empty candidate set or the ones with the most probable predicted entities being either the \textit{None} entity or not in the candidate set, will be automatically ignored.
Thus candidate sets help prevent over-generation and improve the overall quality of entity linking.

\textbf{Entity Disambiguation.} Common methods include: (1) employing constrained beam search to generate the entities with the candidate sets constraining the search process \citep{de2020autoregressive, xiao2023instructed}, (2) \textit{argmax} selection over the predicted probability distributions over the candidate set \citep{broscheit2020investigating, de2021highly, shavarani2023spel}, and (3) ranking the entity candidates using a similarity metric which scores (entity prediction, mention) representation pairs.

For the latter method, the mention representations can be generated by combining the character, word, or token embeddings within the mention span \citep{kolitsas-etal-2018-end,peters2019knowledge,martins-etal-2019-joint,Fevry}, or of the context surrounding the mention \citep{REL_EL, martins-etal-2019-joint}, or all the embeddings of the candidates for the mention \citep{poerner2019bert, feng2022efficient}. The entity representations can come from a pre-computed source like KB+Yago \citep{D17-1277} in \citep{kolitsas-etal-2018-end, REL_EL, peters2019knowledge}, Wikipedia2Vec \cite{Wikipedia2Vec} in \citep{REL_EL, poerner2019bert}, or PPRforNED \citep{pershina-etal-2015-personalized} in \citep{martins-etal-2019-joint}. As well, the entity representations can be textual documents with items like the entity's title, aliases, and description \citep{kannan-ravi-etal-2021-cholan, D11-1072} or model representations of such documents \citep{feng2022efficient, Fevry}. Dot product is most commonly used to score mention-entity similarity \citep{kolitsas-etal-2018-end, peters2019knowledge, poerner2019bert, feng2022efficient, Fevry}. For representations in the form of natural language (as opposed to dense embedding vectors), word overlap, KL-divergence, and n-gram based measures \citep{D11-1072}, as well as document similarity prediction using fine-tuned PLMs \citep{kannan-ravi-etal-2021-cholan} have been used.

Methods using mention-entity similarity tend to also use mention-entity priors (i.e., $\hat{p}(e|m)$) \citep{kolitsas-etal-2018-end, peters2019knowledge, REL_EL, poerner2019bert, feng2022efficient} and thematic coherence (prioritizing entities similar to other entities in the document) \citep{D11-1072, kolitsas-etal-2018-end, REL_EL} to bias the results. The priors are commonly taken from \citet{D17-1277}.

End-to-end entity linking systems usually combine the mention detection and entity disambiguation steps through sharing the same underlying pre-trained language model encoders \cite{kolitsas-etal-2018-end, broscheit2020investigating, shavarani2023spel}. \citet{xiao2023instructed} also combines these steps with their in-context learning approach, which queries a LLM (such as GPT-3, GPT-3.5, or Codex) with a natural language prompt containing the task, candidate entities, and the text to be linked.

\section{Model Changes for Unified Black Box Evaluation}\label{appendix:b}
As mentioned, we use GERBIL to evaluate the models, and \texttt{gerbil\_connect} acts as the middleware between GERBIL and the models. GERBIL provides the model with raw un-tokenized text and expects annotations in the format (\textit{begin char.}, \textit{end char.}, \textit{entity annotation}). In the unification process leading to the specific implementations of \texttt{gerbil\_connect}, we altered each reproduced model to enable them to work within our unified black-box evaluation environment. The following sections discuss each alternation in more details. Table \ref{tab:modifications} discusses the required alterations to adapt each model to our evaluation framework.

We encountered limitations in assessing certain models within our framework. The models from \citet{martins-etal-2019-joint, Fevry, mrini-etal-2022-detection} were not included due to the unavailability of their source code. The model from \citet{kannan-ravi-etal-2021-cholan} was also excluded, as we lacked access to their pre-trained model and the training script was not provided. The models from \citet{broscheit2020investigating, xiao2023instructed} were not considered due to the absence of their pre-trained models and the extended duration of their training process, which was not feasible for inclusion in the manuscript.

For the purpose of facilitating replication and standardizing the assessment of emerging entity linking methodologies in subsequent studies, we strongly advise authors to make their source code, trained models, and a \texttt{gerbil\_connect} integration with GERBIL publicly available. In the absence of these, we suggest the provision of a straightforward function that accepts raw text and outputs a list of annotations. This approach would streamline the process of incorporating an entity linking technique into any evaluation setting.

\begin{table}[ht!]
    \centering
    \addtolength{\tabcolsep}{-0.3em}
    \scalebox{0.95}{
    \begin{tabular}{l|c}
                                         &  Modifications  \\
        \hline\hline
        \citet{kolitsas-etal-2018-end}   & - \\
        
        \citet{peters2019knowledge} & 1, 2, 3 \\
        
        \citet{poerner2019bert}          & 1, 2, 3, 5 \\

        \citet{REL_EL} & - \\
        
        \citet{de2020autoregressive}     & 2, 4 \\
        
        \citet{de2021highly}             & 4 \\
        
        \citet{zhang2021entqa}           & - \\

        \citet{feng2022efficient}        & 1, 2, 3, 5 \\
        
        \citet{shavarani2023spel} & - \\
        \hline\hline
    \end{tabular}
    }
    \caption{Comparison of major modifications made to each system to fit into the unified evaluation environment.}
    \label{tab:modifications}
\end{table}

\subsection{Input Tokenization}
Some models require their input to be tokenized, but do not have this implemented.  \citet{peters2019knowledge} reads directly from the AIDA dataset file where each token is on a new line. Our evaluation environment gives the model text, so we add a tokenization step before the model so the text is closer to the expected input. We used NLTK's word tokenizer for this, as it tokenizes the text similar to the expected CoNLL tokenized format.
After the model makes annotations for each of these word-level tokens, we add another step at the end to convert the predictions from word level to character level.

\subsection{Document Splitting}
Many models have a length limit for the documents they annotate, and therefore need a document splitting strategy.
\citet{de2020autoregressive} uses a document splitting strategy during evaluation \footnote{see \url{https://github.com/facebookresearch/GENRE/issues/30}}, however this strategy is not described in the paper and the code is not in their released repository. We used the splitting strategy created by \citet{bast-etal-2022-elevant} who were able to replicate \citet{de2020autoregressive}.

We also add a step at the end to map the annotations for each section back into the original document.

\subsection{Token-to-Character Annotations}
Many models convert the gold annotations to token-level, and evaluate on their token-level predictions. This allows them to forgo converting their token-level annotations back into character-level annotations. However, our evaluation environment requires character-level annotations as output, so we add a step to perform this conversion where necessary.

\subsection{Outside or Custom Data}
The data used by most models to perform entity linking are available to download. However, \citet{de2020autoregressive} requires custom data that hasn't been released\footnote{see \url{https://github.com/facebookresearch/GENRE/issues/37} and Appendix A.2 Setting in \citet{de2020autoregressive}.}. To fix this, we used the data created by \citet{bast-etal-2022-elevant}. \citet{de2021highly} also requires custom data created by the authors, but this data has been provided in their released repository.

\subsection{Training a New Model}
Although some publications have not released their pre-trained checkpoints, they have released their training data and scripts. In such cases, we trained new model checkpoints to reproduce the original results.

\section{Evaluated Model Descriptions}\label{appendix:c}

Here we provide overviews of the evaluated models in our benchmarking experiments. For brevity, we don't discuss every detail here, and solely rely on the important design choices in each method.

\subsection{Traditional Models}

Traditional models rely on pre-computed dictionaries such as KB+Yago \citep{D17-1277} to provide candidate sets for detected mentions. Replacing the dictionary with a static candidate set for all mentions - such as an empty set or a large set with 5K entities results in poor performance.

\subsubsection{\citet{kolitsas-etal-2018-end}}

For mention detection and candidate generation, \citet{kolitsas-etal-2018-end} use KB+Yago \citep{D17-1277} on all n-length spans in the input text. For entity disambiguation, \citet{kolitsas-etal-2018-end} generate mention representations from the tokens in each mention, then use dot-product similarity to rank each candidate entity using \citet{D17-1277} entity representations. In addition, mention-entity priors and thematic coherence are used to bias each candidate entity's score.

\subsubsection{\citet{peters2019knowledge}}

For mention detection and candidate generation, \citet{peters2019knowledge} use KB+Yago \citep{D17-1277} on all n-length spans in the input text. For entity disambiguation, \citet{peters2019knowledge} generate mention representations from the word pieces in each span, then use dot-product similarity to rank each candidate entity using \citet{D17-1277} entity representations. In addition, mention-entity priors are used to bias each candidate entity's score.

\subsubsection{\citet{REL_EL}}

For mention detection, \citet{REL_EL} use Flair \citep{akbik-etal-2018-contextual} NER annotation tool. Mention representations are generated using their surrounding 50-word context. For candidate generation, \citet{REL_EL} use KB+Yago \cite{D17-1277}. Then, they refine the candidate list by selecting the 4 entities with the highest priors and 3 with the highest dot-product similarity between the mention representation and their \citet{D17-1277} embedding. For entity disambiguation, \citet{REL_EL} attempt to maximize the sum of \textit{context coherence} (a coherence score between an entity and the mention's local context) and thematic coherence for all mentions. The following providing the context coherence calculation formula:
$$\psi(e,c) = \sum_{w \in c} \beta(w) x_w^{\top} B x_{w}$$
where $c$ is context (made up of words $w$), $e$ is the entity, $\beta(w)$ is a weight for each word and $B$ is a trainable matrix.

\subsubsection{\citet{poerner2019bert}}
For mention detection and candidate generation, \citet{poerner2019bert} use KB+Yago \citep{D17-1277} on all n-length spans in the input text. For entity disambiguation, entity candidates are ranked using dot-product similarity. Entity representations are retrieved from Wikipedia2Vec \citep{Wikipedia2Vec}. Mention representations are generated using a BERT model \cite{devlin-etal-2019-bert}. The mention span in the input text is masked by the sum of its candidate entity representations. The mention representation is the mask's embedding generated by the BERT model. In addition, mention-entity priors are used to bias the results.

\subsubsection{\citet{kannan-ravi-etal-2021-cholan}}

For mention detection, each token is classified using a BERT model fine-tuned for BIO tagging. For candidate generation, \citet{kannan-ravi-etal-2021-cholan} use KB+Yago \citep{D17-1277} and the BM25 algorithm \citep{robertson1995okapi} to query a local index \citep{sakor-etal-2019-old} of items from Wikidata. For entity disambiguation, \citet{kannan-ravi-etal-2021-cholan} use a BERT model \textit{WikiBERT} fine-tuned to predict document similarity. Wikipedia articles are used as the entity documents, and the local context of the mention in the text is used as the mention document.

\subsubsection{\citet{feng2022efficient}}
For mention detection and candidate generation, \citet{feng2022efficient} use KB+Yago \citep{D17-1277} on all n-length spans in the input text. For entity disambiguation, entity candidates are ranked using dot-product similarity. Entity representations are generated by encoding the labels of other entities connected through \textit{instance\_of} edges in the KB. Mention representations are generated using a BERT model \cite{devlin-etal-2019-bert}. The mention span in the input text is masked by the sum of its candidate entity representations. The mention representation is the mask's embedding generated by the BERT model. In addition, mention-entity priors are used to bias the results.

\subsection{Generative Models}

Generative models are able to generate candidate sets without relying on a pre-computed mention-to-candidates dictionary. This makes them more robust in settings where these mention-specific candidate sets are lacking. However, some can still use these candidate sets to improve results.

\subsubsection{\citet{de2020autoregressive}}

\citet{de2020autoregressive} output an annotated text of the input by generating it token-by-token. At every token, the model decides either to continue the input or to start a mention. Once inside a mention, the model decides at what token the mention span ends. KB+Yago \citep{D17-1277} is used to obtain a modified candidate set. For entity disambiguation, a candidate is generated using beam search constrained with a trie generated from the candidate entities.

\subsubsection{\citet{de2021highly}}

For mention detection, \citet{de2021highly} classify each token as the start or end of a mention. For candidate generation, \citet{de2021highly} use a given candidate set for each mention. For entity disambiguation, the model scores each candidate with an LSTM then uses a classifier to re-rank the candidates. When not provided with the candidate sets, the model can use the LSTM to generate candidates using constrained beam search.

\subsubsection{\citet{zhang2021entqa}}

For candidate generation, \citet{zhang2021entqa} generate document-level candidates by selecting the top-K entities based on dot-product similarity using an encoder to create an encoding for both the input document and all the entities in Wikipedia (approximately 6M). For mention detection, it classifies each token as a start or end of a mention for each entity. For entity disambiguation, it chooses the entity with the start and end probability scores above a certain threshold.

\subsubsection{\citet{xiao2023instructed}}

For candidate generation, \citet{xiao2023instructed} use dot product similarity to rank entity encodings against the input document encoding, both generated using fine-tuned BERT \cite{devlin-etal-2019-bert} text encoders. For mention detection, it generates a list of possible surface forms for each candidate entity using a modified KB+Yago \citep{D17-1277} dictionary. Then it searches the text for the generated surface forms. For entity disambiguation, it uses constrained beam search to both decide if each candidate mention is valid, and which entity is the correct entity for that mention.

They also experiment with another approach, \textit{INSGENEL-ICL}, which sends a natural language prompt to a LLM with the task description, input document, and candidate entities, and expects a list of annotations as an answer.

\subsection{Structured Prediction Based Models}

These models don't rely on mention-specific candidate sets or need to generate their own. They disambiguate among the entire entity vocabulary. However, some can still use candidate sets to improve results.

\subsubsection{\citet{broscheit2020investigating}}

\citet{broscheit2020investigating} performs entity linking as a per-token classification task over the entire entity vocabulary (the 700K most frequent entities in English Wikipedia). They fine-tune a BERT model for this task.

\subsubsection{\citet{shavarani2023spel}}
\citet{shavarani2023spel} model entity linking as structured prediction with a variable classification vocabulary size. The top-k entities for each token are collected, and merged with adjacent tokens belonging to the same span. When a mention-specific candidate set is available, entities not in this candidate set are filtered out from the mention's candidate entities.

\end{document}

%% file: testa_before_analysis_figure.tex
\begin{tikzpicture}
\begin{axis}[
    width=\columnwidth,
    ybar stacked,   
    bar width=16pt, 
    symbolic x coords={\citet{kolitsas-etal-2018-end}, \citet{peters2019knowledge}, \citet{poerner2019bert}, \citet{REL_EL}, \citet{de2020autoregressive}, \citet{de2021highly}, \citet{zhang2021entqa}, \citet{feng2022efficient}, \citet{shavarani2023spel}},
    xtick=data,
    x tick label style={rotate=45,anchor=east},
    ymin=0,
    ymax=0.4,
    enlarge x limits=0.05,
]

\addplot+[ybar,  draw=gray, fill=gray!30!white, pattern=vertical lines, pattern color=gray] plot coordinates {(\citet{kolitsas-etal-2018-end},0.055520768) (\citet{peters2019knowledge},0.033124053) (\citet{poerner2019bert},0.060530161) (\citet{REL_EL},0.136923398) (\citet{de2020autoregressive},0.057190566) (\citet{de2021highly},0.062250775) (\citet{zhang2021entqa},0.075558339)  (\citet{feng2022efficient},0.056355667) (\citet{shavarani2023spel},0.064913379)};

\addplot+[ybar, draw=red, fill=red!30!white, pattern=horizontal lines, pattern color=red] plot coordinates {(\citet{kolitsas-etal-2018-end},0.058234189) (\citet{peters2019knowledge},0.258930504) (\citet{poerner2019bert},0.034022125) (\citet{REL_EL},0.018159048) (\citet{de2020autoregressive},0.048424129) (\citet{de2021highly},0.029020824) (\citet{zhang2021entqa},0.056564392)  (\citet{feng2022efficient},0.04404091) (\citet{shavarani2023spel},0.050093926)};

\addplot+[ybar, draw=teal, fill=teal!30!white, pattern=north east lines, pattern color=teal] plot coordinates {(\citet{kolitsas-etal-2018-end},0.070757671) (\citet{peters2019knowledge},0.069928556) (\citet{poerner2019bert},0.107493216) (\citet{REL_EL},0.180755583) (\citet{de2020autoregressive},0.094552286) (\citet{de2021highly},0.087727071) (\citet{zhang2021entqa},0.123565018)  (\citet{feng2022efficient},0.109371739) (\citet{shavarani2023spel},0.059277813)};

\addplot+[ybar, draw=blue, fill=blue!30!white, pattern=north west lines, pattern color=blue] plot coordinates {(\citet{kolitsas-etal-2018-end},0.008557712) (\citet{peters2019knowledge},0.018185754) (\citet{poerner2019bert},0.003965769) (\citet{REL_EL},0.006261741) (\citet{de2020autoregressive},0.00667919) (\citet{de2021highly},0.004209127) (\citet{zhang2021entqa},0.024212064)  (\citet{feng2022efficient},0.009392611) (\citet{shavarani2023spel},0.011688583)};

\end{axis}
\end{tikzpicture}

%% file: testa_after_analysis.tex
\begin{tikzpicture}
\begin{axis}[
    width=\columnwidth,
    ybar stacked,
    bar width=25pt,
    symbolic x coords={\citet{poerner2019bert}, \citet{de2020autoregressive}, \citet{de2021highly}, \citet{zhang2021entqa}, \citet{feng2022efficient},  \citet{shavarani2023spel}},
    xtick=data,
    x tick label style={rotate=45,anchor=east},
    ymin=0,
    ymax=1,
    enlarge x limits=0.1,
]

\addplot+[ybar,  draw=gray, fill=gray!30!white, pattern=vertical lines, pattern color=gray] plot coordinates {(\citet{poerner2019bert},0.1903) (\citet{de2020autoregressive},0.09371) (\citet{de2021highly},0.05927) (\citet{zhang2021entqa},0.07555) (\citet{feng2022efficient},0.04779) (\citet{shavarani2023spel},0.06491)};

\addplot+[ybar, draw=red, fill=red!30!white, pattern=horizontal lines, pattern color=red] plot coordinates {(\citet{poerner2019bert},0.03423) (\citet{de2020autoregressive},0.03736) (\citet{de2021highly},0.02943) (\citet{zhang2021entqa},0.05656) (\citet{feng2022efficient},0.2400) (\citet{shavarani2023spel},0.05009)};

\addplot+[ybar, draw=teal, fill=teal!30!white, pattern=north east lines, pattern color=teal] plot coordinates {(\citet{poerner2019bert},0.6988) (\citet{de2020autoregressive},0.1095) (\citet{de2021highly},0.3623) (\citet{zhang2021entqa},0.1235)(\citet{feng2022efficient},0.4362) (\citet{shavarani2023spel},0.05927)};

\addplot+[ybar, draw=blue, fill=blue!30!white, pattern=north west lines, pattern color=blue] plot coordinates {(\citet{poerner2019bert},0.05072) (\citet{de2020autoregressive},0.005009) (\citet{de2021highly},0.003965) (\citet{zhang2021entqa},0.02421) (\citet{feng2022efficient},0.02546) (\citet{shavarani2023spel},0.01168)};

\end{axis}
\end{tikzpicture}

%% file: testa_pr_diff_nocite.tex
\begin{tikzpicture}
\begin{axis}[ 
xbar, xmin=-100, xmax=20,
xlabel={Percentage \%},
symbolic y coords={
    {Shavarani and Sarkar (2023)},
    {Feng et al. (2022)},
    {Zhang et al. (2021)},
    {De Cao et al. (2021)},
    {De Cao et al. (2020)},
    {Poerner et al. (2019)}
},
ytick=data,
nodes near coords, 
nodes near coords align={horizontal},
ytick=data,
width=7cm,
height=7cm,
]
\addplot[pattern=north west lines, pattern color=red] coordinates {
    (2.17,{Shavarani and Sarkar (2023)})
    (-57.86,{Feng et al. (2022)})
    (0,{Zhang et al. (2021)})
    (-23.21,{De Cao et al. (2021)})
    (-5.59,{De Cao et al. (2020)})
    (-66.27,{Poerner et al. (2019)})};
\addplot[pattern=north east lines, pattern color=blue] coordinates {
    (-6.31,{Shavarani and Sarkar (2023)})
    (-48.44,{Feng et al. (2022)})
    (0,{Zhang et al. (2021)})
    (-27.38,{De Cao et al. (2021)})
    (-4.28,{De Cao et al. (2020)})
    (-66.78,{Poerner et al. (2019)})};
\end{axis}
\end{tikzpicture}